\documentclass[conference]{IEEEtran}
\IEEEoverridecommandlockouts
\usepackage{cite}
\usepackage{amsmath,amssymb,amsfonts}
\usepackage{algorithmic}
\usepackage{graphicx}
\usepackage{textcomp}

\newcommand*{\rom}[1]{\expandafter\@slowromancap\romannumeral #1@}
\makeatother
\usepackage{algorithm, algorithmic}
\usepackage{mathtools}
\usepackage{amsmath}
\usepackage{amssymb}
\usepackage{cite}
\usepackage{threeparttable}
\usepackage{calc}
\usepackage[table]{xcolor}
\newcommand{\mycc}{\cellcolor{lightgray}}
\usepackage{textgreek}
\usepackage{booktabs}
\usepackage{siunitx}
\usepackage{array}
\usepackage{balance}

\newcolumntype{L}{>{\phantom{$\mathbin{-}$}$}l<{$}}

\newcolumntype{M}[1]{>{\centering\arraybackslash}m{#1}}
\newcolumntype{N}{@{}m{0pt}@{}}
\usepackage{url}
\usepackage{hyperref}
\usepackage{makecell}
\usepackage{enumitem}
\usepackage{ragged2e} 
\usepackage{multirow}

\usepackage[a4paper, total={184mm,239mm}]{geometry}
\def\BibTeX{{\rm B\kern-.05em{\sc i\kern-.025em b}\kern-.08em
    T\kern-.1667em\lower.7ex\hbox{E}\kern-.125emX}}

\begin{document}

\title{SINA: A Circuit \textbf{S}chematic \textbf{I}mage–to–\textbf{N}etlist Generator Using \textbf{A}rtificial Intelligence}
\author{
\IEEEauthorblockN{
Saoud~Aldowaish, Yashwanth~Karumanchi, Kai-Chen~Chiang, Soroosh~Noorzad, Morteza~Fayazi}
\IEEEauthorblockA{
\textit{University of Utah, Salt Lake City, UT, USA}}
\IEEEauthorblockA{
Emails: \{u1275778, u1518595, kai.chiang, soroosh.noorzad, m.fayazi\}@utah.edu
}}

\maketitle
\begin{abstract}
Current methods for converting circuit schematic images into machine-readable netlists struggle with component recognition and connectivity inference. In this paper, we present SINA, an open-source, fully automated circuit schematic image-to-netlist generator. SINA integrates deep learning for accurate component detection, Connected-Component Labeling (CCL) for precise connectivity extraction, and Optical Character Recognition (OCR) for component reference designator retrieval, while employing a Vision–Language Model (VLM) for reliable reference designator assignments. In our experiments, SINA achieves 96.47\% overall netlist-generation accuracy, which is 2.72x higher than the state-of-the-art approaches.
\end{abstract}


\section{Introduction}
Recent advances in Artificial Intelligence (AI) have reshaped circuit design workflows, especially with the adoption of Large Language Models (LLMs)~\cite{abbineni2025muallm}. Although LLMs have shown strong performance in digital circuit generation~\cite{thakur2024verigen}, their use in analog and mixed-signal design remains limited due to the absence of machine-readable representations of existing circuit knowledge~\cite{bhandari2024masala, fayazi2021applications}. Circuit schematics appearing in research manuscripts, textbooks, and websites capture substantial, validated circuit design information, yet they are not directly interpretable by Electronic Design Automation (EDA) tools. Unlocking this visual information for reuse and for building datasets that support circuit-focused AI models requires converting schematic images into machine-readable netlists~\cite{matsuo2024schemato}.

Current EDA workflows still depend on manually transcribing schematic images into netlists, a process that is slow, error-prone, and difficult to scale~\cite{kelly2023digitizing}. Although several tools have attempted to automate this step, existing image-to-netlist methods continue to struggle with persistent limitations: inaccurate component recognition, unreliable connectivity inference, and weak reference designator extraction and assignment:~\cite{zhang2025analogxpert, fayazi2022fascinet, colter2022tablext}. These limitations hinder the creation of large, high-quality netlist datasets and restrict the capabilities of AI-driven EDA systems.

In this paper, we introduce SINA, an open-source~\cite{SINA} and fully automated pipeline for converting schematic images into SPICE-compatible netlists. SINA is designed to operate robustly across diverse schematic styles. To address the key shortcomings of existing methods, SINA integrates a YOLO-based~\cite{redmon2016you} detector for component recognition, applies Connected-Component Labeling (CCL)~\cite{opencv_ccl} for reliable connectivity inference, and combines Optical Character Recognition (OCR) with a Vision–Language Model (VLM) for accurate reference designator extraction and assignment.

\begin{figure*}[t]
\centering
\includegraphics[width=0.90\textwidth]{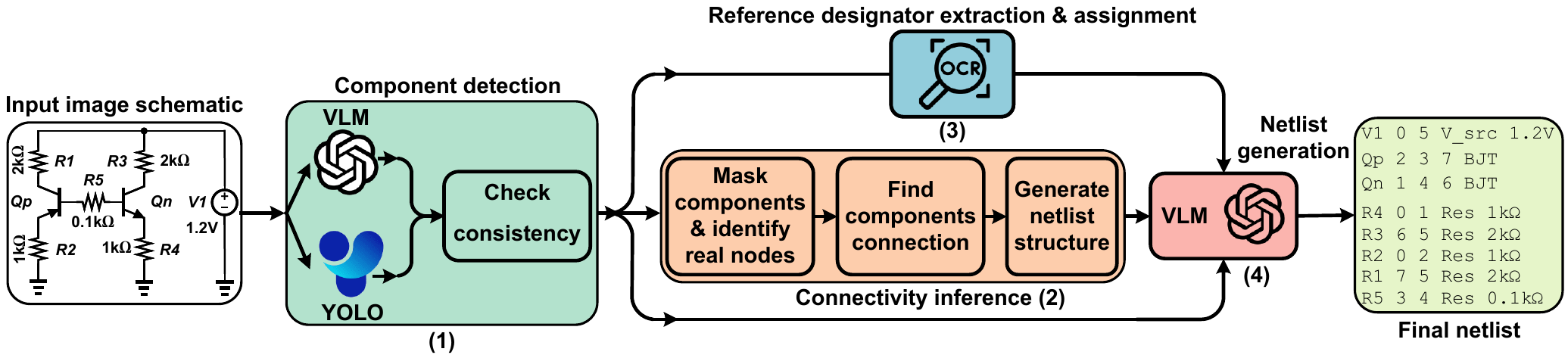}
\caption{The proposed SINA's workflow.}
\label{fig:overall_workflow}
\end{figure*}

\begin{figure*}[t]
    \centering
    \includegraphics[width=0.95\textwidth]{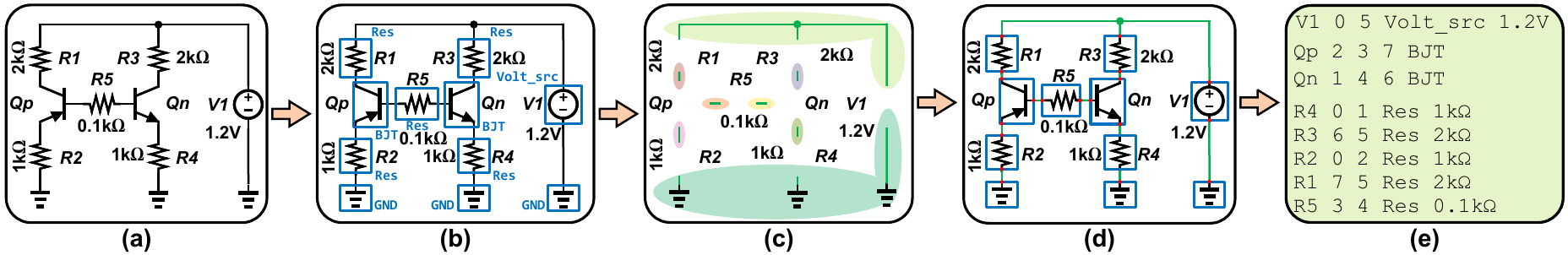}
    \caption{An example of SINA's pipeline. (a) The original circuit schematic image. (b) The detection model output with identified components and their bounding boxes. (c) Removing components and clustering nodes. (d) Nodes and their connections to the components. (e) The final generated netlist.}
    \label{fig:ic_example}
\end{figure*}

\section{Proposed Framework}
SINA is a fully automated pipeline that takes circuit schematic images as input and converts them into SPICE-compatible netlists. As illustrated in Fig.~\ref{fig:overall_workflow}, the framework consists of four main stages: component detection, connectivity inference, reference designator (label) extraction \& assignment, and final netlist generation.

\subsection{Component Detection}
SINA uses a YOLOv11-based object detection model~\cite{redmon2016you} to identify circuit components, returning bounding boxes, component types, and unique component identifiers, as shown in Fig.~\ref{fig:ic_example}(b). To enhance reliability, SINA includes an independent VLM-based verification step using GPT-4o~\cite{openai2023gpt4}. The VLM analyzes the schematic. Its predictions of component types and counts are compared against YOLO’s detections to compute a concordance-based confidence score. Any mismatches between the two systems are flagged for users review.

\subsection{Connectivity Inference}
SINA infers electrical connectivity by determining which component terminals share a common circuit node. After component detection, the detected components are masked using their bounding boxes, leaving only the wiring of the circuit visible. The resulting image is processed using CCL~\cite{opencv_ccl}, which segments the wiring into distinct connected regions. Regions that correspond to artifacts such as small stubs, loops, or gaps are filtered out, and only those that connect to two or more components are retained. This is consistent with the definition of an electrical net~\cite{gray2009analysis}. Electrically equivalent nodes (\textit{e.g.} multiple ground symbols) are merged into a single node, as shown in Fig.~\ref{fig:ic_example}(c). Component-to-node relationships are then established by identifying intersection points between component positions and node regions, determining which terminals connect to which nodes, as illustrated in Fig.~\ref{fig:ic_example}(d). This stage produces the complete component-to-node mapping required for netlist generation.

\subsection{Reference Designator Extraction and Assignment}
SINA employs EasyOCR~\cite{jaided2020easyocr} to extract textual annotations from the schematic, enabling the identification of reference designators and component values. The OCR-extracted text is mapped to detected components based on spatial proximity, providing essential context for the next stage, where GPT-4o assigns final component designators and values. By separating text extraction from semantic interpretation, SINA leverages each tool’s strengths: the OCR for dependable text detection and VLMs for contextual reasoning.

\subsection{Netlist Generation}
As shown in Fig.~\ref{fig:ic_example}(e), a VLM (GPT-4o) generates the SPICE-compatible netlist using three inputs: the OCR-extracted reference designators, the component-to-node connectivity mappings, and the original schematic image for visual context. With this information, the model assigns final reference designators and values to each component, and produces the final netlist that reflects the circuit’s schematic structure and parameters.

\begin{table}[t]
\begin{center}
\begin{threeparttable}
\centering
\caption{SINA's component detection performance evaluation.}
\def\arraystretch{1.2}\tabcolsep 2pt
\label{table:component_detection_performance_evaluation}
\begin{tabular}{|M{12mm}|M{12mm}|M{12mm}|M{45mm}|}
\hline\hline
Precision & Recall & F1 Score & Weighted Mean Average Precision\\
\hline
94\% & 99\% & 96.47\% & 98\%\\
\hline
\end{tabular}
\end{threeparttable}
\end{center}
\end{table}

\begin{table}[t]
\begin{center}
\begin{threeparttable}
\centering
\caption{Performance comparison between SINA and Masala-CHAI~\cite{bhandari2024masala} on 40 test schematics.}
\def\arraystretch{1.1}
\label{table:sina_masalachai_comparison}
\begin{tabular}{|M{11mm}|M{12mm}|M{23mm}|M{11mm}|M{11mm}|}
\hline\hline
Method & Text Extraction & Component Detection  F1 Score & Circuit Structure & Overall Accuracy\\
\hline
\mycc SINA & \mycc \textbf{97.55\%} & \mycc \textbf{99.6\%} & \mycc \textbf{99.3\%} & \mycc \textbf{96.47\%} \\
\hline
\cite{bhandari2024masala} & 95.09\% & 62.4\% & 59.8\% & 35.5\% \\
\hline
\end{tabular}
\end{threeparttable}
\end{center}
\end{table}

\section{Evaluation}
To evaluate the performance of SINA, we construct an open-source benchmark containing schematics that span a broad range of styles, complexities, and component counts. Our object detection model is fine-tuned on a custom dataset of more than 700 annotated schematics, drawn from diverse sources such as research manuscripts, scanned textbooks, and hand-drawn sketches. To increase robustness across varied visual conditions, we apply data augmentation techniques including scaling, brightness jitter, and flipping~\cite{shorten2019survey}.

We assess the component detection model performance using a benchmark of 75 schematics containing more than 1,000 components across 10 distinct types. The benchmark includes a mix of computer-generated, scanned, and hand-drawn schematics, covering a broad range of styles and component layouts. The component detection model achieves an F1 score of 96.47\%, as shown in Table~\ref{table:component_detection_performance_evaluation}.

For overall netlist generation performance, we compare SINA with Masala-CHAI~\cite{bhandari2024masala}, the only available open-source netlist generation framework for this task, using a curated set of 40 schematics. These test circuits are selected to ensure a fair comparison, including only component types supported by both systems. As shown in Table~\ref{table:sina_masalachai_comparison}, SINA outperforms Masala-CHAI across all evaluation metrics, with an overall accuracy that is \textbf{2.72x} higher. 

\section{Conclusion}
This paper introduced SINA, a fully automated and open-source pipeline for converting circuit schematic images into SPICE-compatible netlists. SINA combines a YOLOv11-based model for component detection, CCL for connectivity inference, OCR for text extraction, and a VLM for resolving reference designator assignments. In our evaluation, SINA achieves 96.47\% netlist-generation accuracy, representing a 2.72x improvement compared to the state-of-the-art approaches.

\newpage
\newpage
\bibliographystyle{IEEEtran}

\begin{thebibliography}{10}
\providecommand{\url}[1]{#1}
\csname url@samestyle\endcsname
\providecommand{\newblock}{\relax}
\providecommand{\bibinfo}[2]{#2}
\providecommand{\BIBentrySTDinterwordspacing}{\spaceskip=0pt\relax}
\providecommand{\BIBentryALTinterwordstretchfactor}{4}
\providecommand{\BIBentryALTinterwordspacing}{\spaceskip=\fontdimen2\font plus
\BIBentryALTinterwordstretchfactor\fontdimen3\font minus \fontdimen4\font\relax}
\providecommand{\BIBforeignlanguage}[2]{{%
\expandafter\ifx\csname l@#1\endcsname\relax
\typeout{** WARNING: IEEEtran.bst: No hyphenation pattern has been}%
\typeout{** loaded for the language `#1'. Using the pattern for}%
\typeout{** the default language instead.}%
\else
\language=\csname l@#1\endcsname
\fi
#2}}
\providecommand{\BIBdecl}{\relax}
\BIBdecl

\bibitem{abbineni2025muallm}
P.~Abbineni, S.~Aldowaish, C.~Liechty, S.~Noorzad, A.~Ghazizadeh, and M.~Fayazi, ``MuaLLM: A Multimodal Large Language Model Agent for Circuit Design Assistance with Hybrid Contextual Retrieval-Augmented Generation,'' \emph{arXiv preprint arXiv:2508.08137}, 2025.

\bibitem{thakur2024verigen}
S.~Thakur \emph{et~al.}, ``VeriGen: A Large Language Model for Verilog Code Generation,'' \emph{ACM Transactions on Design Automation of Electronic Systems}, 2024.

\bibitem{bhandari2024masala}
\BIBentryALTinterwordspacing
J.~Bhandari, V.~Bhat, Y.~He, H.~Rahmani, S.~Garg, and R.~Karri, ``Masala-CHAI: A Large-Scale Spice Netlist Dataset for Analog Circuits by Harnessing AI,'' 2025. [Online]. Available: \url{https://arxiv.org/abs/2411.14299}
\BIBentrySTDinterwordspacing

\bibitem{fayazi2021applications}
M.~Fayazi, Z.~Colter, E.~Afshari, and R.~Dreslinski, ``Applications of Artificial Intelligence on The Modeling and Optimization for Analog and Mixed-Signal Circuits: A Review,'' \emph{IEEE Transactions on Circuits and Systems I: Regular Papers}, vol.~68, no.~6, pp. 2418--2431, 2021.

\bibitem{matsuo2024schemato}
R.~Matsuo \emph{et~al.}, ``Schemato -- An LLM for Netlist-to-Schematic Conversion,'' \emph{arXiv preprint arXiv:2411.13899}, 2024.

\bibitem{kelly2023digitizing}
J.~Kelly and M.~Cole, ``Digitizing images of electrical-circuit schematics,'' in \emph{Proc. ACM Symposium on Document Engineering}, 2023.

\bibitem{zhang2025analogxpert}
H.~Zhang, S.~Sun, Y.~Lin, R.~Wang, and J.~Bian, ``AnalogXpert: Automating Analog Topology Synthesis by Incorporating Circuit Design Expertise into Large Language Models,'' in \emph{2025 International Symposium of Electronics Design Automation (ISEDA)}.\hskip 1em plus 0.5em minus 0.4em\relax IEEE, 2025, pp. 772--777.

\bibitem{fayazi2022fascinet}
M.~Fayazi \emph{et~al.}, ``FASCINET: A Fully Automated Single-Board Computer Generator Using Neural Networks,'' \emph{IEEE Transactions on Computer-Aided Design of Integrated Circuits and Systems}, vol.~41, no.~12, pp. 5435--5448, 2022.

\bibitem{colter2022tablext}
Z.~Colter \emph{et~al.}, ``Tablext: A combined neural network and heuristic based table extractor,'' \emph{Array}, vol.~15, p. 100220, 2022.

\bibitem{SINA}
S.~Aldowaish \emph{et~al.}, ``SINA - Schematic Image to Netlist Tool,'' \url{https://anonymous.4open.science/r/SINA-213F/README.md}, 2026.

\bibitem{redmon2016you}
J.~Redmon, S.~Divvala, R.~Girshick, and A.~Farhadi, ``You Only Look Once: Unified, Real-Time Object Detection,'' in \emph{Proc. CVPR}, 2016, pp. 779--788.

\bibitem{opencv_ccl}
OpenCV, ``Connected Components with Stats — OpenCV Documentation,'' \url{https://docs.opencv.org/4.x/d3/dc0/group__imgproc__shape.html#gaffe776513b24d84b39af8ab0930fef7f}, 2023.

\bibitem{openai2023gpt4}
OpenAI, ``GPT-4 Technical Report,'' \url{https://arxiv.org/abs/2303.08774}, 2023, arXiv:2303.08774.

\bibitem{gray2009analysis}
P.~R. Gray, P.~J. Hurst, S.~H. Lewis, and R.~G. Meyer, \emph{Analysis and Design of Analog Integrated Circuits}.\hskip 1em plus 0.5em minus 0.4em\relax John Wiley \& Sons, 2009.

\bibitem{jaided2020easyocr}
JaidedAI, ``EasyOCR: Ready-to-use OCR with 80+ Languages Supported,'' \url{https://github.com/JaidedAI/EasyOCR}, 2020.

\bibitem{shorten2019survey}
C.~Shorten and T.~M. Khoshgoftaar, ``A Survey on Image Data Augmentation for Deep Learning,'' \emph{Journal of Big Data}, vol.~6, no.~1, pp. 1--48, 2019.

\end{thebibliography}

\end{document}